\documentclass{article}
\usepackage{ijcai22}

\usepackage{times}
\usepackage{soul}
\usepackage{url}
\usepackage[hidelinks]{hyperref}
\usepackage[utf8]{inputenc}
\usepackage[small]{caption}
\usepackage{graphicx}
\usepackage{amsmath}
\usepackage{amsthm}
\usepackage{booktabs}
\usepackage{algorithm}
\usepackage{algorithmic}
\usepackage{booktabs}
\usepackage{multirow}
\usepackage{amssymb}
\usepackage{amsmath}        
\usepackage{graphicx}       
\usepackage{subfigure}      

\title{Self-Supervised Information Bottleneck for Deep Multi-View Subspace Clustering}

\author{
    Shiye Wang, Changsheng Li, Yanming Li, Ye Yuan, Guoren Wang
    \affiliations
    School of Computer Science and Technology, Beijing Institute of Technology, Beijing, China
    \emails
    \{sywang,lcs,lym,yuan-ye\}@bit.edu.cn, wanggrbit@126.com
}

\begin{document}

\maketitle

\begin{abstract}
In this paper, we explore the problem of deep multi-view subspace clustering framework from an information-theoretic point of view. We extend the traditional information bottleneck principle to learn common information among different views in a self-supervised manner, and accordingly establish a new framework called  Self-supervised Information Bottleneck based Multi-view Subspace Clustering  (SIB-MSC). Inheriting the advantages from information bottleneck, SIB-MSC can learn a latent space for each view to capture common information among the latent representations of different views by removing superfluous information from the view itself while retaining sufficient information for the latent representations of other views. 
Actually, the latent representation of each view provides a kind of self-supervised signal for training the latent representations of other views. Moreover, SIB-MSC  attempts to learn the other latent space for each view to capture the view-specific information by introducing mutual information based regularization terms, so as to further improve the performance of multi-view subspace clustering.
To the best of our knowledge, this is the first work to explore information bottleneck for multi-view subspace clustering.
Extensive experiments on real-world multi-view data demonstrate that our method achieves superior performance over the related state-of-the-art methods.
\end{abstract}

\section{Introduction}
Multi-view subspace clustering aims to discover the underlying subspace structures of  data by fusing different view information. A popular strategy is to take advantage of a linear model to learn an affinity matrix for each view, and then build a consensus graph based on these affinity matrices
~\cite{gunnemann2012multi}.
Following this line, many approaches have been proposed in the past decade.
For instance,  \cite{gao2015multi} performs subspace clustering on each view, and uses a common cluster structure to make sure the consistence among different views. \cite{cao2015} employs the Hilbert Schmidt Independence Criterion (HSIC) \cite{gretton2005measuring} to explore the complementary information of different views, and obtains diverse subspace representations.
\cite{zhang2015low} presents a low-rank tensor constraint to explore the complementary information from multiple views, which can capture the high order correlations of multi-view data. Furthermore, \cite{zhang2017latent} attempts to seek a latent representation to well reconstruct the data of all views, and perform clustering based on the learned latent representation.
\cite{kang2020large} proposes to learn a smaller graph between the raw data points and the generated anchors for each view, and then designs an integration mechanism to merge those graphs, making the eigen-value decomposition accelerated significantly. 
However, all of these methods use linear models to model multi-view data, thus often failing in handling data with complex (often nonlinear) structures.

In recent years, with the development of deep learning techniques, a few deep learning based methods have been gradually proposed for multi-view subspace clustering \cite{abavisani2018deep,zhu2019multi,DBLP:journals/pami/ZhangFHCXTX20}, which mainly aim to leverage deep learning to non-linearly map data into latent spaces for capturing complementary information or diverse information among different views. 
The representative works including Deep Multimodal Subspace  Clustering  networks (DMSC) \cite{abavisani2018deep}, Multi-View Deep Subspace Clustering Networks (MvDSCN) \cite{zhu2019multi}, generalized Latent Multi-View Subspace Clustering (gLMSC) \cite{DBLP:journals/pami/ZhangFHCXTX20}. 
Because of the powerful representation ability of deep learning, these methods achieve promising performance.

Recently, information bottleneck has attracted increasing attention in the machine learning community \cite{motiian2016information,DBLP:conf/iclr/Federici0FKA20,wan2021multi,DBLP:conf/iclr/YuXRBHH21}. Information bottleneck is an information theoretic principle that can learn a minimal sufficient representation for downstream prediction tasks \cite{tishby2000information}. 
So far, information bottleneck has been widely applied to representation learning \cite{wan2021multi}, graph learning \cite{DBLP:conf/nips/WuRLL20}, 
visual recognition \cite{motiian2016information}, etc. However, there is few work to explore information bottleneck for deep multi-view subspace clustering.

\begin{figure}[t]
\centering
\includegraphics[width=0.7\columnwidth]{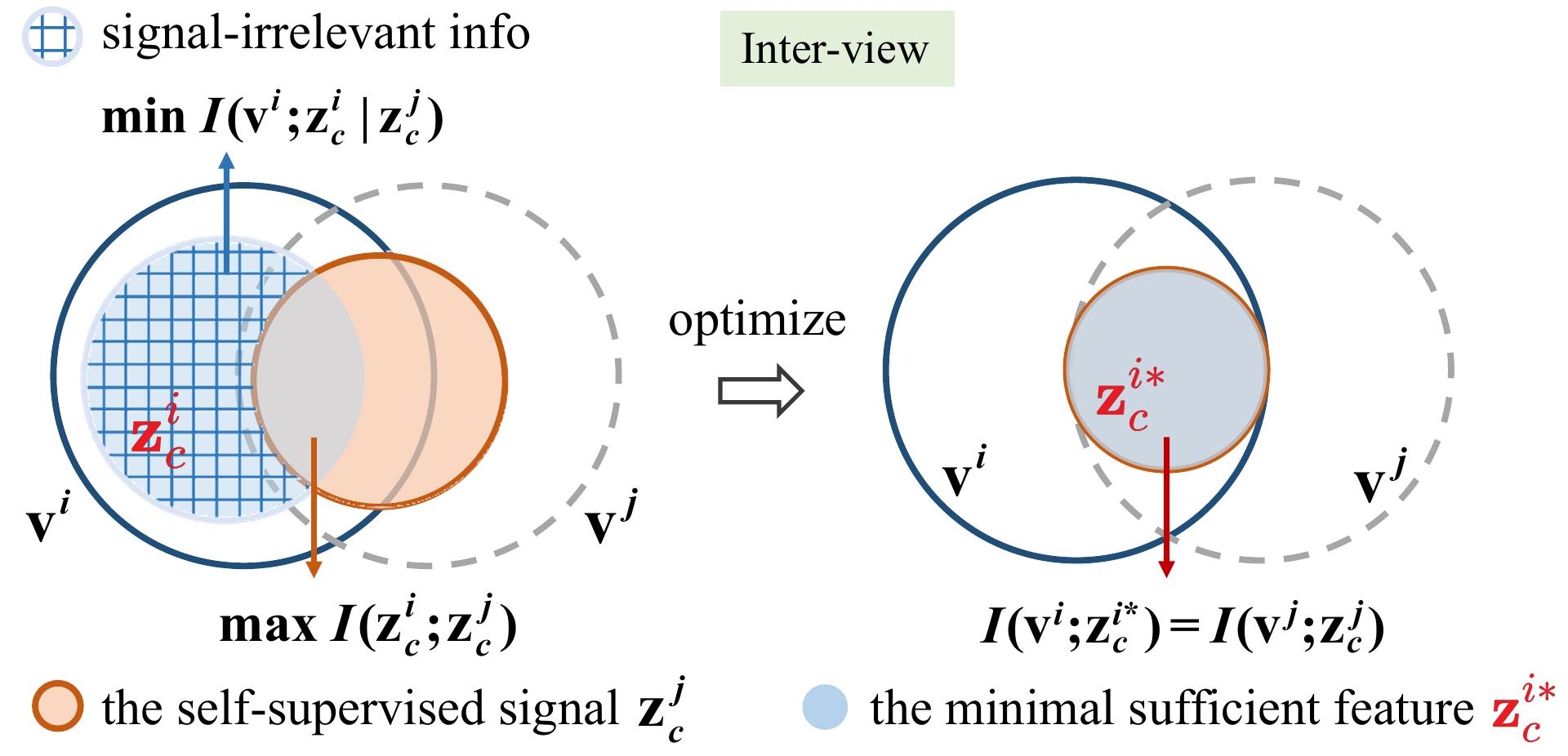} 
\vspace{-0.05in}
\caption{High-level takeaways for our self-supervised information bottleneck.}
\label{fig1}
\vspace{-0.15in}
\end{figure}

In light of this, we propose a new deep multi-view subspace clustering framework in this paper, called  Self-supervised Information Bottleneck based Multi-view Subspace Clustering (SIB-MSC). In order to learn the view-common information, we extend information bottleneck to learn the latent space of each view by removing  superfluous  information  from  the  view  itself  while  retaining sufficient information for the latent representations of other views. 
As shown in Figure \ref{fig1}, taking the latent representation $\mathbf{z}_c^j$ extracted from the $j$-th view as an anchor, we expect that $\mathbf{z}_c^i$ can remove superﬂuous information from the original input $\mathbf{v}^i$ of the $i$-th view  by minimizing the conditional mutual information $I(\mathbf{v}^i;\mathbf{z}_c^i|\mathbf{z}_c^j)$ while preserving sufficient information for $\mathbf{z}_c^j$ by maximizing the mutual information between $\mathbf{z}_c^i$ and $\mathbf{z}_c^j$.
Actually, the latent representation $\mathbf{z}_c^j$ serves as a self-supervised signal to guide the learning of the latent representation $\mathbf{z}_c^i$. Moreover, we attempt to learn view-specific information for each view by imposing mutual information based constraints to  further improve the representation ability of the model. After that, the affinity matrices based on the minimal sufficient view-common representations is taken as the input to spectral clustering.

In summary, our contributions are four-fold:
\begin{itemize}
\item We propose an information bottleneck based framework for deep multi-view subspace clustering. To the best of our knowledge, this is the first work to explore information bottleneck for multi-view subspace clustering. 
\item We put forward to learn the minimal sufficient latent representation for each view {with the guidance  of self-supervised information bottleneck, which can obtain common information among different views}. 
\item {We present mutual information based constraints to capture a view-specific space} for each view to be complementary to the view-common space, and  well reconstruct the samples, such that the performance can be further improved.

\item
Extensive experiments on multi-view data verify the effectiveness of our proposed model. 
Moreover, our model achieves superior performance over the existing deep multi-view subspace clustering algorithms.
\end{itemize}

\section{Related Work}
In this section, we review some related works on deep multi-view subspace clustering and information bottleneck based representation learning.
\paragraph{Deep Multi-View Subspace Clustering.}
Deep learning has demonstrated the powerful representation ability in various learning tasks. 
In recent years, there are a few works which attempt to leverage deep learning to solve the problem of multi-view subspace clustering \cite{abavisani2018deep,zhu2019multi,DBLP:journals/pami/ZhangFHCXTX20}.
DMSC \cite{abavisani2018deep} takes advantage of deep learning to investigate the early, intermediate and late fusion strategies to learn a latent space which is used to discover the subspace structures of the data.
\cite{zhu2019multi} proposes to learn a view-specific self-representation matrix for each view and a common self-representation matrix for all views. Then, the learned common self-representation matrix is used to perform spectral clustering \cite{ng2002spectral}. 
gLMSC \cite{DBLP:journals/pami/ZhangFHCXTX20} aims to integrate multi-view inputs into a comprehensive latent representation which can encode complementary information from different views and well capture 
the underlying subspace structure. 
Different from these works, we focus on studying the problem of deep multi-view subspace clustering from an information-theoretic point of view. 

\paragraph{Representation Learning with Information Bottleneck.}
The information bottleneck principle  can learn a robust representation by removing information from the input that is not relevant to a given task.   \cite{tishby2000information}. 
Because of its effectiveness, information bottleneck has been successfully applied to representation learning \cite{motiian2016information,DBLP:conf/iclr/AlemiFD017,DBLP:conf/iclr/Federici0FKA20,wan2021multi,DBLP:conf/iclr/YuXRBHH21}.
\cite{motiian2016information} extends the information bottleneck principle to leverage an auxiliary data view during training for learning a better visual classiﬁer.
VIB ~\cite{DBLP:conf/iclr/AlemiFD017} is a variational approximation to the information bottleneck, which can parameterize the information bottleneck method using a neural network. 
MIB~\cite{DBLP:conf/iclr/Federici0FKA20} is proposed to minimize the mutual information between two different views to reduce the redundant information across them, where the original input is regarded as a self-supervised signal.
\cite{wan2021multi} utilizes the information
bottleneck principle to remove the superfluous information from
the multi-view data. 
\cite{DBLP:conf/iclr/YuXRBHH21} applies information bottleneck to identify the maximally informative yet compressive subgraph for subgraph recognition. 
As aforementioned, we attempt to extend the information bottleneck principle to deep multi-view subspace clustering, motivated by these methods.

\section{The Proposed Method}
\begin{figure*}[t] 
\centering
\includegraphics[width=0.8\textwidth]{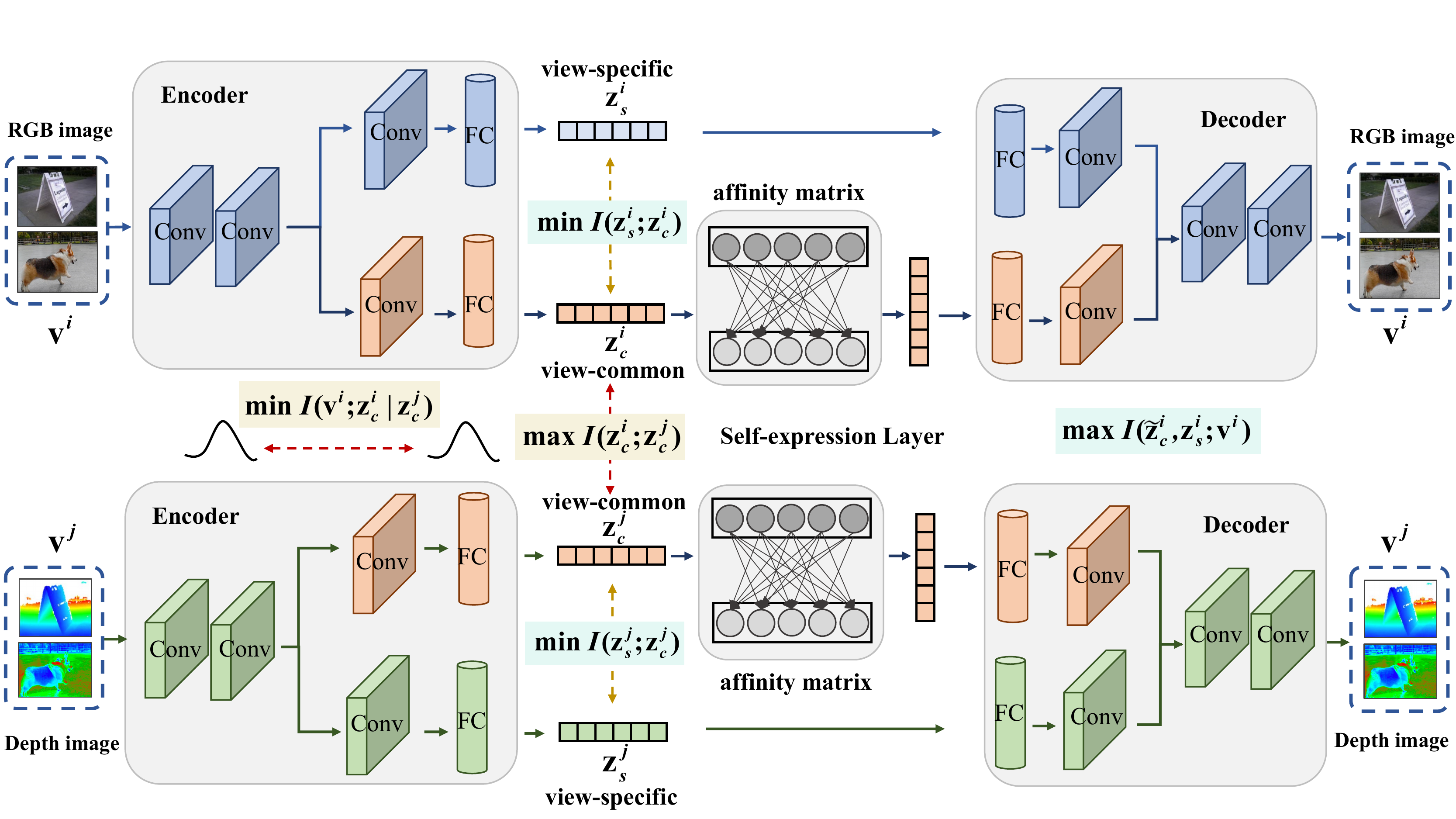} 
\caption{Illustration of the overall architecture. SIB-MSC extends information bottleneck to learn a latent space $\mathbf{z}_c^{i}$ for each the $i-$th view  for capturing view-common information among different views.
Taking the latent representation $\mathbf{z}_c^j$ as an anchor, we expect to eliminate the useless information from the original view $i$ by minimizing $I(\mathbf{v}^i;\mathbf{z}_c^i|\mathbf{z}_c^j)$ while preserving sufficient information for $\mathbf{z}_c^j$ by maximizing $I(\mathbf{z}_c^i;\mathbf{z}_c^j)$. 
Here, the latent representation $\mathbf{z}_c^j$ plays a role of self-supervision.
{Moreover, the view-specific feature $\mathbf{z}_s^{i}$ is extracted to maintain the  complementary information, by  minimizing $I(\mathbf{z}_s^i;\mathbf{z}_c^i)$ to { disentangle} from $\mathbf{z}_c^i$ and maximizing $I(\widetilde{\mathbf{z}}_c^i,\mathbf{z}_s^i;\mathbf{v}^i)$  to achieve { jointly informative} to the original inputs,} such that the performance of multi-view subspace learning can be further improved.
After that, the affinity matrices obtained based on the view-common representations are taken as the input to spectral clustering. 
}
\label{fig2}
\vspace{-0.1in}
\end{figure*}

In this section, we will elaborate the details of the proposed deep multi-view subspace clustering framework, as shown in Figure 2.
For better clarification, we first give some notations.

\paragraph{Notation Statement.}
Suppose that we have $m$ views, let
$\mathbf{V}=\{\mathbf{V}^{1},\cdots, \mathbf{V}^{m}\}$ be the input samples, where 
$\mathbf{V}^{i}=\{ \mathbf{v}_1^{i}, \mathbf{v}_2^{i},\cdots , \mathbf{v}_n^{i} \} \in \mathbb{R}^{n \times d_{i}}$ 
denotes the samples of the $i$-th view, $d_{i}$ denotes the feature dimension of the $i$-th view, and $n$ is the number of samples. $\mathbf{v}^{i}_k$ denotes the $k$-th sample of view $i$. For writing conveniently, we use the notation $\mathbf{v}^i$ to take place of $\mathbf{v}^{i}_k$ throughout the paper. 
For each sample $\mathbf{v}^{i}$, we extract two kinds of features: a view-common feature $\mathbf{z}_{c}^{i} \in \mathbb{R}^{d}$ and a view-specific feature $\mathbf{z}_{s}^{i} \in \mathbb{R}^{d}$. 
Given three random variables, ${X}$, ${Y}$ and ${Z}$, we measure the mutual dependence between $X$ and $Y$ with the aid of mutual information $I({X};{Y})$. 
In addition, we can leverage conditional mutual information $I({X};{Y}|{Z})$ to measure the amount of information that $Y$ can capture from $X$ that is irrelevant to $Z$. 

\subsection{Learning View-Common via Information Bottleneck}
To capture the view-common information, we extend information bottleneck to reach this goal, where  the latent representation of each view is learnt by discarding useless information  from  the  view  that is not relevant to other views. 
In this paper, we first consider the case of two views, in order to simplify the statement. It is easily extended to the general case which has more than two views.

As shown in Figure \ref{fig2},  to make the learnt $\mathbf{z}_c^i$ and $\mathbf{z}_c^j$ contain the common information as much as possible, we take $\mathbf{z}_c^i$ and $\mathbf{z}_c^j$ as self-supervised signals to guide the learning of each other, respectively. Here, we take $\mathbf{z}_c^j$ as an example to show how to guide the learning of $\mathbf{z}_c^i$. The learning process of $\mathbf{z}_c^j$ is the same. We first decompose the mutual information between $ \mathbf{v}^{i} $ and $ \mathbf{z}_{c}^{i} $ into two components:
\begin{align}\label{eq1}
I(\mathbf{v}^{i};\mathbf{z}_{c}^{i})=\underbrace{I(\mathbf{v}^{i};\mathbf{z}_{c}^{i}|\mathbf{z}_{c}^{j})}_{\rm signal-irrelevant} + \underbrace{I(\mathbf{z}_{c}^{i};\mathbf{z}_{c}^{j})}_{\rm signal-relevant},
\end{align}
where the conditional mutual information $I(\mathbf{v}^{i};\mathbf{z}_{c}^{i}|\mathbf{z}_{c}^{j})$ {represents the information that $\mathbf{z}_{c}^{i}$ can capture from $\mathbf{v}^{i}$ and is not relevant to $\mathbf{z}_{c}^{j}$}, i.e., the superfluous information.  $I(\mathbf{z}_{c}^{i};\mathbf{z}_{c}^{j})$ measures how much information of $\mathbf{z}_{c}^{j}$ is accessible from $\mathbf{z}_{c}^{i}$.

In Eq. (\ref{eq1}), we minimize the signal-irrelevant part to discard the superﬂuous information, and maximize the signal-relevant part  to encourage $\mathbf{z}_{c}^{i}$ to be sufficient for $\mathbf{z}_{c}^{j}$. Therefore, we propose to minimize the following loss function to learn  the minimal sufficient feature $\mathbf{z}_{c}^{i}$ as:
\begin{align}\label{loss1}
\mathcal{L}_{c}^{i} =- I(\mathbf{z}_{c}^{i};\mathbf{z}_{c}^{j}) + \lambda_{i} I(\mathbf{v}^{i};\mathbf{z}_{c}^{i}|\mathbf{z}_{c}^{j}), 
\end{align} where $\lambda_{i}$ is the Lagrangian multiplier introduced by the information constraint.

Since it is difficult to directly handle the conditional mutual information, we can conduct the transformation based on the chain rule of the conditional mutual information as:
\begin{align} \label{eq3}
&I(\mathbf{v}^{i};\mathbf{z}_{c}^{i}|\mathbf{z}_{c}^{j})=I(\mathbf{v}^{i};\mathbf{z}_{c}^{i},\mathbf{z}_{c}^{j})-I(\mathbf{v}^{i};\mathbf{z}_{c}^{j}) \nonumber  \\
&=I(\mathbf{v}^{i};\mathbf{z}_{c}^{i})+I(\mathbf{z}_{c}^{j};\mathbf{v}^{i}|\mathbf{z}_{c}^{i}) -I(\mathbf{v}^{i};\mathbf{z}_{c}^{j}) \nonumber  \\
&=I(\mathbf{v}^{i};\mathbf{z}_{c}^{i})-I(\mathbf{v}^{i};\mathbf{z}_{c}^{j}).
\end{align}  


Based on Eq. (\ref{eq3}), we can rewrite Eq. (\ref{loss1}) as:
\begin{align}
\mathcal{L}_{c}^{i} &=- I(\mathbf{z}_{c}^{i};\mathbf{z}_{c}^{j}) + \lambda_{i} I(\mathbf{v}^{i};\mathbf{z}_{c}^{i}|\mathbf{z}_{c}^{j}) \nonumber  \\
=- &I(\mathbf{z}_{c}^{i};\mathbf{z}_{c}^{j}) + \lambda_{i}( I(\mathbf{v}^{i};\mathbf{z}_{c}^{i})-I(\mathbf{z}_{c}^{j};\mathbf{v}^{i})).
\end{align} 


Let us first study the cross-view mutual information maximization $I(\mathbf{z}_{c}^{j};\mathbf{v}^{i})$. Considering the fact that $\mathbf{v}^{i}$ is over-informative for subspace clustering concerning more about the structure information, we relieve $\mathbf{v}^{j}$ to $\mathbf{h}^{j}$ which is the exact feature the encoder extracted before splitting into two parts, thus we have $I(\mathbf{z}_{c}^{j};\mathbf{v}^{i}) \geq I(\mathbf{z}_{c}^{j};\mathbf{h}^{i}) $.


Jointly considering the $i$-th and $j$-th views, we can obtain the loss function for learning the view-common latent space:
\begin{align}\label{losscommon}
\mathcal{L}_{c} = &\mathcal{L}_{c}^{i} +\mathcal{L}_{c}^{j} \nonumber \\
=&- 2*I(\mathbf{z}_{c}^{i};\mathbf{z}_{c}^{j}) + \lambda_{i} (I(\mathbf{v}^{i};\mathbf{z}_{c}^{i})-I(\mathbf{z}_{c}^{j};\mathbf{v}^{i})) \nonumber \\
&+\lambda_{j} (I(\mathbf{v}^{j};\mathbf{z}_{c}^{j})-I(\mathbf{z}_{c}^{i};\mathbf{v}^{j}))  \nonumber \\
\leq &- 2*I(\mathbf{z}_{c}^{i};\mathbf{z}_{c}^{j}) -(\lambda_{i} I(\mathbf{z}_{c}^{j};\mathbf{h}^{i})+ \lambda_{j} I(\mathbf{z}_{c}^{i};\mathbf{h}^{j})) \nonumber \\
&+ (\lambda_{i} I(\mathbf{v}^{i};\mathbf{z}_{c}^{i}) + \lambda_{j} I(\mathbf{v}^{j};\mathbf{z}_{c}^{j})).
\end{align} 
where we can divide the above loss function into three types of terms  as $\mathcal{L}_{c}^{dis}=-2*I(\mathbf{z}_{c}^{i};\mathbf{z}_{c}^{j})$,
$\mathcal{L}_{c}^{cmi}=-(\lambda_{i} I(\mathbf{z}_{c}^{j};\mathbf{h}^{i})+\lambda_{j}I(\mathbf{z}_{c}^{i};\mathbf{h}^{j}))$,
$\mathcal{L}_{c}^{mkl}=\lambda_{i} I(\mathbf{v}^{i};\mathbf{z}_{c}^{i}) + \lambda_{j} I(\mathbf{v}^{j};\mathbf{z}_{c}^{j})$. 
We denote these notations, with the purpose of performing the ablation study in the experiment conveniently.


Thus, minimizing $\mathcal{L}_c$ can be relaxed to minimize its upper bound, i.e., the right part of the inequality sign.
Since it is difficult to directly optimize Eq. (\ref{losscommon}) due to involving several mutual information terms, we attempt to seek an approximate solution to Eq. (\ref{losscommon}). 
Here we only present our formulation for view $i$, as the one for view $j$ is analogous.
We first consider the cross-view mutual information maximization:
\begin{align}\label{trans}
I(\mathbf{z}_{c}^{j};\mathbf{h}^{i})= D_{KL}[p(\mathbf{z}_{c}^{j},\mathbf{h}^{i})||p(\mathbf{z}_{c}^{j})p(\mathbf{h}^{i})].
\end{align}

As the essence of mutual information is KL divergence and the maximization of KL-divergence is divergent, we instead seek for the convergent Jensen-Shannon MI estimator as in ~\cite{nowozin2016f}, thus a tractable estimation of $I(\mathbf{z}_{c}^{j};\mathbf{h}^{i})$ can be defined as:
\begin{align}
\max \quad &\mathbb{E}_{(\mathbf{z}_{c}^{j};\mathbf{h}^{i})}[\log \mathrm{D}(\mathbf{z}_{c}^{j},\mathbf{h}^{i})]  \nonumber \\
&+ \mathbb{E}_{(\mathbf{z}_{c}^{j};\mathbf{\hat{h}}^{i})}[\log (1-\mathrm{D}(\mathbf{z}_{c}^{j},\mathbf{\hat{h}}^{i}))],
\end{align}
where the feature $\mathbf{z}_{c}^{j}$ serves as the anchor, and the features $\mathbf{z}_{c}^{j}$ and $\mathbf{h}^{i}$ originated from view $i$ of the same example are regarded as a  positive pair. $\mathbf{\hat{h}}^{i}$ denotes random sampled negative input from view $i$, and $\mathrm{D}$ indicates the discriminator to estimate the probability of the input pair.

Similarly, the mutual information maximization term $I(\mathbf{z}_{c}^{i};\mathbf{z}_{c}^{j})$ can  be also tractable by replacing $\mathbf{h}^{i}$ wth $\mathbf{z}_{c}^{i}$ in Eq. (\ref{trans}).

As for the mutual information minimization term $I(\mathbf{v}^{i};\mathbf{z}_{c}^{i})$, let $q(\mathbf{z}_{c}^{i})$ be a variational approximation of $p(\mathbf{z}_{c}^{i})$, with $D_{KL}[p(\mathbf{z}_{c}^{i}),q(\mathbf{z}_{c}^{i})] \geq 0 \Rightarrow \int p(\mathbf{z}_{c}^{i})\log p(\mathbf{z}_{c}^{i})d\mathbf{z}_{c}^{i}  \geq  \int p(\mathbf{z}_{c}^{i})\log  q(\mathbf{z}_{c}^{i})d\mathbf{z}_{c}^{i}$, we can deduce a variational upper bound as follow:
\begin{align}
&I(\mathbf{v}^{i};\mathbf{z}_{c}^{i}) =  \mathbb{E}_{p(\mathbf{v}^{i})}[D_{KL}(p(\mathbf{z}_{c}^{i}|\mathbf{v}^{i})|p(\mathbf{z}_{c}^{i})] \nonumber \\
&=  \mathbb{E}_{p(\mathbf{v}^{i})}[D_{KL}(p(\mathbf{z}_{c}^{i}|\mathbf{v}^{i})|q(\mathbf{z}_{c}^{i})]   
 -\mathbb{E}_{p(\mathbf{z}_{c}^{i})}[D_{KL}(p(\mathbf{z}_{c}^{i})|q(\mathbf{z}_{c}^{i})]  \nonumber \\ 
&\leq \mathbb{E}_{p(\mathbf{v}^{i})}[D_{KL}(p(\mathbf{z}_{c}^{i}|\mathbf{v}^{i})|q(\mathbf{z}_{c}^{i})],
\end{align}

The above inequation enforces the latent representation $\mathbf{z}_{c}^{i}$ conditioned on $\mathbf{v}^{i}$ to a predefined distribution $q(\mathbf{z}_{c}^{i})$ such as a standard Gaussian distribution. 
Due to space limitations, we put the above derivations in the supplementary material.

\subsection{Learning View-Specific via Mutual Information}
As aforementioned, we aim to learn the other latent space for each view to capture view-specific information. We expect that the view-specific latent space $\mathbf{z}_{s}^{i}$ is complementary to the view-common representation $\mathbf{z}_{c}^{i}$.
Thus we propose to minimize the following mutual information based loss function:
\begin{align}\label{lossspecific}
\mathcal{L}_{s} = \mathcal{L}_{s}^{i} + \mathcal{L}_{s}^{j}
=I(\mathbf{z}_{s}^{i};\mathbf{z}_{c}^{i}) 
+I(\mathbf{z}_{s}^{j};\mathbf{z}_{c}^{j}) 
\end{align}

Inspired by the predictability minimization model \cite{6795705}, we present a strategy to optimize the above objective function. 
Taking $I(\mathbf{z}_{s}^{i};\mathbf{z}_{c}^{i})$ as an example, we first design a predictor $F$ for one feature to anticipate another feature by calculate $\mathbf{z}_{c}^{i}$ conditioned on $\mathbf{z}_{s}^{i}$, and if $\mathbf{z}_{c}^{i}$ is conditional independent of $\mathbf{z}_{s}^{i}$, then it is unpredictable. Thus the corresponding objective is:
\begin{align}
\min \max [\mathbb{E}_{p(\mathbf{z}_{s}^{i}|\mathbf{v}^{i})}[F(\mathbf{z}_{c}^{i}|\mathbf{z}_{s}^{i})] + \mathbb{E}_{p(\mathbf{z}_{c}^{i}|\mathbf{v}^{i})}[F(\mathbf{z}_{s}^{i}|\mathbf{z}_{c}^{i})]]
\end{align}
where the predictor $F$ targets at properly making anticipation while the encoder $p(\mathbf{z}_{s}^{i}|\mathbf{v}^{i})$ and $p(\mathbf{z}_{c}^{i}|\mathbf{v}^{i})$ make the prediction process hard. 

\subsection{Learning Self-Expressiveness  via View-Common}
In order to discover the subspace structures of the data, we introduce a self-expressive layer based on the view-common space, motivated by the work in \cite{NIPS2017_e369853d}.

Specifically, we introduce a fully-connected linear layer to learn self-expressiveness based on the view-common features, and minimize the  following loss function for  view $i$:
\begin{align}\label{selfres}
\mathcal{L}_{se}^{i} = &||\mathbf{Z}_{c}^{i} - \mathbf{Z}_{c}^{i}\mathbf{C}^{i}||_{F}^{2} + \lambda||\mathbf{C}^{i}||^{2}   \nonumber \\
&{\rm s.t.}\; \text{diag} (\mathbf{C}^{i})=0 ,
\end{align}
where $\mathbf{C}^{i}$ is the weights of the self-expressive layer for view $i$, which  serves as the affinity matrix for spectral clustering. And $\lambda \geq 0$ is one trade-off parameter.

\subsection{Reconstruction via Mutual  Information}
To improve the representation ability of the model, we introduce a mutual information based regularization term to reconstruct the samples.
For writing conveniently, we use the notation $\widetilde{\mathbf{Z}}_{c}^{i}$ to replace $\mathbf{Z}_{c}^{i}\mathbf{C}^{i}$ in the rest of the paper. 
Let $\widetilde{\mathbf{z}}_{c}^{i}$ denotes the $k$-th sample of $\widetilde{\mathbf{Z}}_{c}^{i}$ .
By maximizing the mutual information $I(\mathbf{v}^{i};\mathbf{z}_{s}^{i},\widetilde{\mathbf{z}}_{c}^{i})$ between the concatenated feature $(\mathbf{z}_{s}^{i},\widetilde{\mathbf{z}}_{c}^{i})$ and the original input, the raw observations can be well reconstructed via the view-common and view-specific features.

Since the multivariate mutual information term is still intractable, we introduce $q(\mathbf{v}^{i}|\mathbf{z}_{s}^{i},\widetilde{\mathbf{z}}_{c}^{i})$ as a variational approximaiton of $p(\mathbf{v}^{i}|\mathbf{z}_{s}^{i},\widetilde{\mathbf{z}}_{c}^{i})$. Then, we have:
\begin{align}\label{lossrec}
&\mathcal{L}_{r}^{i} = I(\mathbf{v}^{i};\mathbf{z}_{s}^{i},\widetilde{\mathbf{z}}_{c}^{i}) \nonumber  \\
&=\mathbb{E}_{p(\mathbf{v}^{i},\mathbf{z}_{s}^{i},\widetilde{\mathbf{z}}_{c}^{i})}\log (q(\mathbf{v}^{i}|\mathbf{z}_{s}^{i},\widetilde{\mathbf{z}}_{c}^{i}))
+ H(\mathbf{v}^{i})  \nonumber  \\
&\quad +\mathbb{E}_{p(\mathbf{z}_{s}^{i},\widetilde{\mathbf{z}}_{c}^{i})}[D_{KL}[p(\mathbf{v}^{i}|\mathbf{z}_{s}^{i},\widetilde{\mathbf{z}}_{c}^{i}))||q(\mathbf{v}^{i}|\mathbf{z}_{s}^{i},\widetilde{\mathbf{z}}_{c}^{i})]]   \nonumber  \\
&\geq \mathbb{E}_{p(\mathbf{v}^{i},\mathbf{z}_{s}^{i},\widetilde{\mathbf{z}}_{c}^{i})}\log (q(\mathbf{v}^{i}|\mathbf{z}_{s}^{i},\widetilde{\mathbf{z}}_{c}^{i})). 
\end{align}

Leveraging Markov chain $\mathbf{z}_{s}^{i} \leftrightarrow \mathbf{v}^{i} \leftrightarrow \widetilde{\mathbf{z}}_{c}^{i}$, we have
\begin{align} \label{self}
I(\mathbf{v}^{i};\mathbf{z}_{s}^{i}&,\widetilde{\mathbf{z}}_{c}^{i}) \geq \mathbb{E}_{p(\mathbf{v}^{i},\mathbf{z}_{s}^{i},\widetilde{\mathbf{z}}_{c}^{i})}\log (q(\mathbf{v}^{i}|\mathbf{z}_{s}^{i},\widetilde{\mathbf{z}}_{c}^{i}))  \nonumber  \\
=& \mathbb{E}_{p(\mathbf{v}^{i})p(\mathbf{z}_{s}^{i}|\mathbf{v}^{i})p(\widetilde{\mathbf{z}}_{c}^{i}|\mathbf{v}^{i})}\log(q(\mathbf{v}^{i}|\mathbf{z}_{s}^{i},\widetilde{\mathbf{z}}_{c}^{i})).
\end{align}

When  $q(\mathbf{v}^{i}|\mathbf{z}_{s}^{i},\widetilde{\mathbf{z}}_{c}^{i})$ is subordinated to the Gaussian distribution, the practical implementation of
$-\mathbb{E}_{p(\mathbf{z}_{s}^{i}|\mathbf{v}^{i})}\mathbb{E}_{p(\widetilde{\mathbf{z}}_{c}^{i}|\mathbf{v}^{i})}\log(q(\mathbf{v}^{i}|\mathbf{z}_{s}^{i},\widetilde{\mathbf{z}}_{c}^{i}))$ is the reconstruction loss as $||\mathbf{v}^{i}-R(\mathbf{z}_{s}^{i},\widetilde{\mathbf{z}}_{c}^{i})||^{2}$, where $R$ is the decoder.

Thus, the final reconstruction loss can be obtained by:
\begin{align}\label{recontruction}
\mathcal{L}_{r} = \mathcal{L}_{r}^{i} + \mathcal{L}_{r}^{j}.
\end{align}

\begin{table*}[t]
\setlength\tabcolsep{3pt}
\centering
\small
\begin{tabular}{@{}cc|ccc|ccccccc|c@{}}
\toprule
Dataset                                                                   & Metric  & LRR   & DSCN  & DPSC  & LMVSC & SiMVC & CoMVC & DMSC  & MvDSCN & gLMSC & MIB-DSC & SIB-MSC         \\ \midrule
\multirow{4}{*}{RGB-D}                                                    & ACC     & 30.0 & 33.9 & 36.4 & 31.0 & 34.4 & 36.6 & 35.4 & 38.8  & 37.6 & 42.4   & \textbf{51.2} \\
                                                                          & NMI     & 58.9 & 58.9 & 59.9 & 54.4 & 60.3 & 61.9 & 60.8 & 63.9  & 61.0 & 65.6   & \textbf{71.1} \\
                                                                          & ARI     & 14.4 & 16.3 & 16.3 & 12.2 & 16.5 & 18.5 & 19.0 & 21.0  & 17.2 & 23.4   & \textbf{32.9} \\
 \midrule
\multirow{3}{*}{\begin{tabular}[c]{@{}c@{}}Fashion\\ -MNIST\end{tabular}} & ACC     & 56.1 & 53.6 & 56.0 & 60.0 & 62.2 & 67.4 & 60.2 & 63.3  & 63.8 & 65.6   & \textbf{72.5} \\
                                                                          & NMI     & 61.9 & 59.4 & 60.6 & 60.4 & 60.8 & 62.6 & 57.1 & 57.4  & 63.8 & 63.6   & \textbf{65.2} \\
                                                                          & ARI     & 41.7 & 40.6 & 42.6 & 44.6 & 45.8 & 50.1 & 44.5 & 45.5  & 45.5 & 49.0   & \textbf{55.5} \\ \midrule
\multirow{3}{*}{\begin{tabular}[c]{@{}c@{}}not\\ -MNIST\end{tabular}}     & ACC     & 45.4 & 48.8 & 47.1 & 49.2 & 51.5 & 57.2 & 51.8 & 48.9  & 51.2 & 55.6   & \textbf{61.9} \\
                                                                          & NMI     & 45.1 & 44.0 & 43.0 & 45.4 & 47.1 & 47.9 & 46.4 & 44.9  & 52.1 & 45.3   & \textbf{55.8} \\
                                                                          & ARI     & 15.8 & 31.8 & 30.3 & 31.8 & 35.6 & 36.3 & 30.1 & 28.4  & 24.5 & 35.6   & \textbf{42.7} \\ \midrule
\multirow{3}{*}{COIL20}                                                   & ACC     & 55.1      & 58.9 & 63.2    & 68.3    & 70.1 & 73.2 & 67.8  & 71.2   & 71.3 & 74.1   & \textbf{78.4} \\
                                                                          & NMI     & 61.8      & 63.2 & 72.1  & 76.5       & 78.8 & 80.7 & 77.3      & 79.1     & 80.7 & 80.3   & \textbf{84.1} \\
                                                                          & ARI     & 37.9      & 44.4 & 50.9  & 53.7      & 60.9 & 65.7 & 58.3       & 63.8      & 53.6 & 67.3   & \textbf{72.8} \\ \bottomrule
\end{tabular}
\vspace{-0.05in}
\caption{Clustering results (\%) on the RGB-D, Fashion-MNIST, notMNIST, and COIL20 datasets.}
\vspace{-0.15in}
\label{general}
\end{table*}

\subsection{Overall Model}
After introducing all the components of this work, we now give the final loss function based on (\ref{losscommon}), (\ref{lossspecific}), (\ref{recontruction}), (\ref{selfres}) as:
\begin{align}
\min \mathcal{L} = \mathcal{L}_{c} + \alpha \mathcal{L}_{s} - \beta \mathcal{L}_{r} + \gamma \mathcal{L}_{se}
\label{total_loss}
\end{align}  
where $\alpha$, $\beta$ and $\gamma$ are three trade-off parameters.


After obtaining $C^{i}$ for each view $i$, we first average all self-expressive coefficient matrices as the final self-expressive coefficient matrix. After that, we  perform subspace clustering, like most of the existing works \cite{zhang2017latent}.
\section{Experiments}

\subsection{Datasets}
We evaluate our method on four publicly available datasets, including one multi-modal dataset RGB-D object dataset \cite{lai2011large}, three image datasets Fashion-MNIST~\cite{xiao2017fashion}, notMNIST ~\cite{bulatov2011notmnist}, and COIL20 \cite{Nene96columbiaobject}. 
For RGB-D,  we use the same data with that of the paper \cite{zhu2019multi}, to conduct a fair comparison.
For Fashion-MNIST and notMNIST, we use the same data with  \cite{li2021lrsc}.
Similar to gLMSC \cite{DBLP:journals/pami/ZhangFHCXTX20}, we extract two types of deep features as two views of these two datasets.
For the COIL20 dataset, we extract the intensity, LBP and Gabor features as three different views.

\subsection{Experimental Setting}
\paragraph{Compared Methods.}
We compare SIB-MSC with the following related methods: three deep multi-view subspace clustering methods including DMSC \cite{abavisani2018deep}, MvDSCN \cite{zhu2019multi},  gLMSC \cite{DBLP:journals/pami/ZhangFHCXTX20}, 
one linear multi-view subspace clustering  \cite{kang2020large},
three single-view subspace clustering including LRR \cite{liu2012robust}, DSCN \cite{NIPS2017_e369853d}, DPSC \cite{zhou2019latent}, 
one information bottleneck based multi-view learning method   MIB-DSC \cite{DBLP:conf/iclr/Federici0FKA20}, one multi-view clustering methods named SiMVC and CoMVC proposed very recently \cite{trosten2021reconsidering}.
Note that MIB is originally designed for multi-view representation learning via information bottleneck. We construct MIB-DSC as our baseline by applying MIB to our view-common feature learning module to obtain the affinity matrix. 
{Single-view clustering methods are applied to each view, and the best performance is reported.} 

\paragraph{Evaluation Metrics and Experimental Protocol.}
For all quantitative evaluations, we use  three popular metrics to evaluate the clustering performance, including {ACC} (Accuracy), {NMI} (Normalized Mutual Information), and {ARI} (Adjusted Rand Index).
For RGBD dataset, we employ a CNN with 3 convolutional layers with [64, 32, 16] channels as the encoder.  For Fashion-MNIST and notMNIST, we employ a CNN with 4 convolutional layers with [40, 30, 20, 10] as the encoder.
For COIL20, four fully connected layers are used as the encoder.
The symmetric network structures are used as the decoder, correspondingly.
For more detailed settings, we report them in the supplementary material.

\begin{figure*}
\centering
\subfigure[v1]{\includegraphics[width=0.18\linewidth]{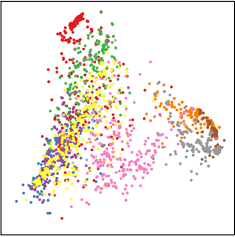}}  %
\subfigure[v2]{\includegraphics[width=0.18\linewidth]{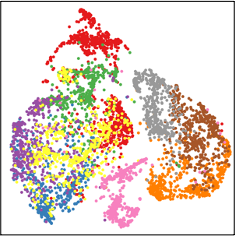}}
\subfigure[Result of v1]{\includegraphics[width=0.18\linewidth]{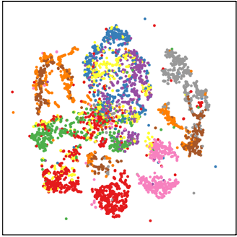}}
\subfigure[Result of v2]{\includegraphics[width=0.18\linewidth]{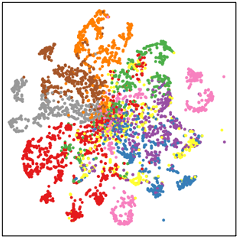}}
\subfigure[Fused result]{\includegraphics[width=0.18\linewidth]{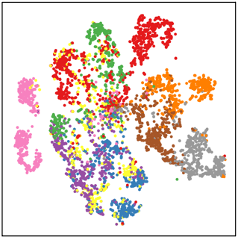}}
\vspace{-0.1in}
\caption{t-SNE visualization of original input and self-expressive matrices obtained by different strategy on the Fashion-MNIST dataset.} 
\label{illustration}
\vspace{-0.05in}
\end{figure*}
\subsection{Expreimental Result}
\paragraph{General Performance.} 
We perform the experiments on the four datasets, and report the results in Table \ref{general}.
Our method consistently outperforms all other competitors. Particularly, when compared with gLMSC, our method outperforms it in terms of all the three metrics.
 This may be because that gLMSC aims to capture the comprehensive information among all views. However, not all the information in each view is helpful for discovering the subspace structures of the data.
 Thus, it is necessary to discard superfluous information while preserve useful information for boosting the performance of subspace clustering.
 Moreover, our method achieves better performance than MIB-DSC in terms of the three evaluation metrics on all the datasets. This demonstrates that utilizing the latent representation of one view as a self-supervised signal to guide the latent representation learning of other view is beneficial to subspace clustering.
In addition, when compared with multi-view clustering methods, SiMVC and CoMVC, our method still obtains superior performance, illustrating that discovering the subspace structure of the data is important for clustering in many real-world applications.

\paragraph{ Effect of Different Sizes of Data.}
We test the performance of the subspace clustering methods using different sizes of data on the Fashion-MNIST dataset.
The results are listed in Table \ref{volume}.
We can seet that our proposed method still outperforms other approaches when varying the sizes of the data.
This further verifies the effectiveness of our method.
Note that we do not perform gLMSC when the number of sample is 10,000, due to its huge memory requirement. 
\begin{table}
\setlength\tabcolsep{3pt}
\centering
\small
\begin{tabular}{@{}c|lll|lll|lll@{}}
\toprule
No. Points & \multicolumn{3}{c|}{1000} & \multicolumn{3}{c|}{5000} & \multicolumn{3}{c}{10000} \\ \midrule
Metric  & ACC       & NMI       & ARI      & ACC       & NMI       & ARI      & ACC       & NMI       & ARI      \\ \midrule
LRR     & 56.1     & 61.9     & 41.7    & 55.1     & 62.5     & 41.5    & 51.9     & 62.5     & 39.9    \\
DSCN    & 53.6     & 59.4     & 40.6    & 56.3     & 55.8     & 37.2    & 58.5     & 62.6     & 44.7    \\ 
DPSC    & 56.0     & 60.6     & 42.6    & 59.9     & 58.4     & 42.5    & 60.4     & 61.5     & 45.4    \\ \midrule
LMVSC   & 60.0     & 60.4     & 44.6    & 64.9     & 64.1     & 50.9    & 61.8     & 62.4     & 47.9    \\
SiMVC   & 62.2     & 60.8     & 45.8    & 67.8     & 65.6     & 54.9    & 68.2     & 66.9     & 54.3    \\
CoMVC   & 67.4     & 62.6     & 50.1    & 72.7     & \textbf{67.0}  & 56.8 & 72.9  & 67.5     & 57.9    \\ 
DMSC    & 60.2     & 57.1     & 44.5    & 58.3     & 59.4     & 43.6    & 56.4     & 60.1     & 42.0    \\
MvDSCN  & 63.3     & 57.4     & 44.5    & 67.1     & 60.6     & 51.2    & 63.2     & 58.8     & 47.6    \\
gLMSC   & 63.8     & 63.8     & 45.5    & 61.3     & 64.8     & 47.4    & $-$       & $-$       & $-$      \\
MIB-DSC & 65.6     & 63.6     & 49.0    & 69.0     & 65.4     & 54.8    & 67.1     & 61.9     & 49.6    \\ \midrule
SIB-MSC  & \textbf{72.5} & \textbf{65.2}  & \textbf{55.5}    & \textbf{74.7}     & 66.1     & \textbf{58.7}    & \textbf{73.4}     & \textbf{69.4}     & \textbf{58.4}    \\ \bottomrule
\end{tabular}
\caption{Clustering results (\%) on the Fashion-MNIST dataset with different sizes of data.}
\label{volume}
\vspace{-0.15in}
\end{table}



\paragraph{Ablation Study.}
We conduct the ablation study on the all four datasets. 
Considering the overall objective $\mathcal{L}$ in (\ref{total_loss}), since the self-expressiveness loss $\mathcal{L}_{self}$ and the reconstruction loss $\mathcal{L}_{r}$ are necessary for subspace clustering, we perform the experiments by removing one of the rest loss terms each time.
We first test whether the view-specific feature is useful for multi-view subspace clustering, and set $\alpha=0$. Then, we test the effectiveness of the loss term $\mathcal{L}_{c}$ which consists of three losses as shown in (\ref{losscommon}). The experimental results are reported in Table \ref{ablation}.
Each component in our method is helpful for multi-view subspace clustering.
\begin{figure}
\centering
\subfigure{\includegraphics[width=0.22\linewidth]{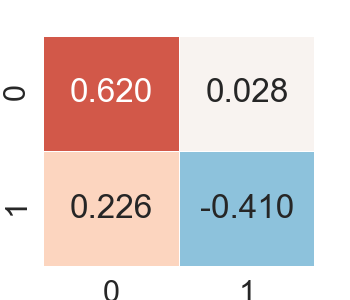}}
\subfigure{\includegraphics[width=0.22\linewidth]{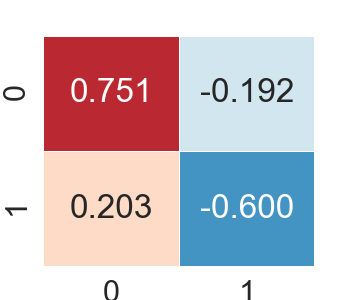}}
\subfigure{\includegraphics[width=0.22\linewidth]{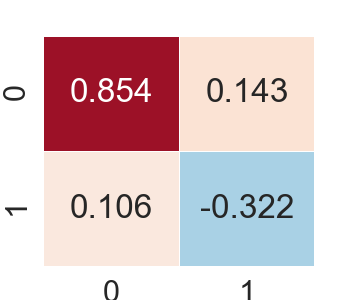}} 
\subfigure{\includegraphics[width=0.25\linewidth]{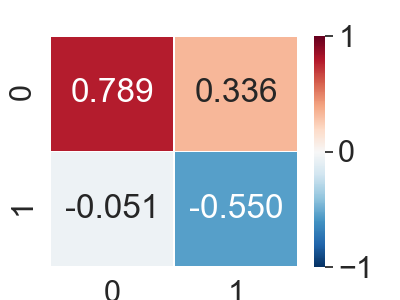}} 
\vspace{-0.05in}
\caption{Correlations of different kinds of features.  The left upper corner denotes the correlation between two view-common features $\mathbf{z}_{c}^{i}$ and $\mathbf{z}_{c}^{j}$. The right bottom corner denotes the correlation  between two view-specific features $\mathbf{z}_{s}^{i}$ and $\mathbf{z}_{s}^{j}$. The left bottom measures the correlation between $\mathbf{z}_{s}^{j}$ and $\mathbf{z}_{c}^{j}$. The right upper corner is the correlation between $\mathbf{z}_{c}^{i}$ and $\mathbf{z}_{s}^{i}$.}
\label{illustration2}
\vspace{-0.1in}
\end{figure}

\paragraph{Visualization.}
To further understand  the proposed model, we provide a visualization of the affinity matrix on the Fashion-MNIST dataset. The results are shown in Figure \ref{illustration}. Figure \ref{illustration}(c) and Figure \ref{illustration}(d) show the results of one of the two views respectively, while Figure \ref{illustration}(e) gives the  result of the fused affinity matrix. We can clearly observe that the fused affinity matrix possesses better representation ability.

We also test the correlations between the view-common features and view-specific features  using the cosine similarity on the Fashion-MNIST.
We randomly select 4 samples and use their latent representations $\mathbf{z}$ to calculate the similarities.
The results are shown in Figure \ref{illustration2}. We can see that $\mathbf{z}_{c}^{i}$ and $\mathbf{z}_{c}^{j}$ have a relatively high correlation, while  $\mathbf{z}_{c}^{i}$ and $\mathbf{z}_{s}^{i}$, $\mathbf{z}_{s}^{j}$ and $\mathbf{z}_{c}^{j}$ have relatively low correlations. This verifies the effectiveness of our proposed method. 

\begin{table}[]
\setlength\tabcolsep{5pt}
\centering
\small
\begin{tabular}{@{}cccccccc@{}}
\toprule                                                                    & $\mathcal{L}_{s}$ & $\mathcal{L}_{c}^{mkl}$ & $\mathcal{L}_{c}^{cmi}$ & $\mathcal{L}_{c}^{dis}$ & ACC   & NMI   & ARI   \\ \midrule
\multirow{5}{*}{\rotatebox{90}{RGBD}}                                                     & $-$    &\checkmark    &\checkmark      &\checkmark      & 43.8 & 66.5 & 24.5 \\
                                                                          &\checkmark      & $-$    &\checkmark      &\checkmark      & 41.8 & 65.3 & 22.7 \\
                                                                          &\checkmark      &\checkmark      & $-$    &\checkmark      & 29.2 & 55.2 & 11.6 \\
                                                                          &\checkmark      &\checkmark      &\checkmark      & $-$    & 48.1 & 69.3 & 28.9 \\
                                                                          &\checkmark      &\checkmark      &\checkmark      &\checkmark      & \textbf{51.2} & \textbf{71.1} & \textbf{31.7} \\ \midrule
\multirow{5}{*}{\begin{tabular}[c]{@{}c@{}}\rotatebox{90}{Fashion}\end{tabular}} & $-$    &\checkmark      &\checkmark      &\checkmark      & 68.5 & 64.9 & 51.6 \\
                                                                          &\checkmark      & $-$    &\checkmark      &\checkmark      & 64.9 & 61.5 & 48.5 \\
                                                                          &\checkmark      &\checkmark      & $-$    &\checkmark      & 63.0 & 55.8 & 43.1 \\
                                                                          &\checkmark      &\checkmark      &\checkmark      & $-$    & 68.3 & 64.6 & 50.9 \\
                                                                          &\checkmark      &\checkmark      &\checkmark      &\checkmark      & \textbf{72.5} & \textbf{65.2} & \textbf{55.5} \\ \midrule
\multirow{5}{*}{\rotatebox{90}{notMNIST}}                                                     & $-$    &\checkmark    &\checkmark      &\checkmark      & 55.0 & 48.1 & 36.1 \\
                                                                          &\checkmark      & $-$    &\checkmark      &\checkmark      & 59.3 & 54.2 & 38.5 \\
                                                                          &\checkmark      &\checkmark      & $-$    &\checkmark      & 58.4 & 52.1 & 42.0 \\
                                                                          &\checkmark      &\checkmark      &\checkmark      & $-$    & 59.0 & 52.5 & 39.5 \\
                                                                          &\checkmark      &\checkmark      &\checkmark      &\checkmark      & \textbf{61.9} & \textbf{55.8} & \textbf{42.7} \\ \midrule
\multirow{5}{*}{\rotatebox{90}{COIL20}}                                                     & $-$    &\checkmark    &\checkmark      &\checkmark      & 77.8 & 82.1 & 70.3 \\
                                                                          &\checkmark      & $-$    &\checkmark      &\checkmark      & 73.2 & 82.2 & 69.2 \\
                                                                          &\checkmark      &\checkmark      & $-$    &\checkmark      & 51.4 & 64.5 & 39.2 \\
                                                                          &\checkmark      &\checkmark      &\checkmark      & $-$    & 68.8 & 75.7 & 59.7 \\
                                                                          &\checkmark      &\checkmark      &\checkmark      &\checkmark      & \textbf{78.4} & \textbf{84.1} & \textbf{72.8} \\ \bottomrule
\end{tabular}
\caption{Ablation study on all the four datasets.}
\label{ablation}
\vspace{-0.15in}
\end{table}

\section{Conclusion}
In this paper, we proposed a new framework for multi-view deep subspace clustering from an information-theoretic point of view. 
We extended information bottleneck to learn the view-common information, where the latent representation of each view served as a self-supervised signal to capture the common information.
Moreover, we attempted to capture the view-specific feature via mutual information to further boost the  model performance.  
Experimental results on four publicly available datasets verified  the effectiveness of our model.

\appendix

\bibliographystyle{named}
\bibliography{ijcai22}

\end{document}